\documentclass{article}

\usepackage{arxiv}

\usepackage[utf8]{inputenc} 
\usepackage[T1]{fontenc}    
\usepackage{hyperref}       
\usepackage{url}            
\usepackage{booktabs}       
\usepackage{amsfonts}       
\usepackage{nicefrac}       
\usepackage{microtype}      
\usepackage{lipsum}		
\usepackage{graphicx}
\usepackage{natbib}
\usepackage{doi}

\usepackage{amsmath}
\usepackage{graphicx,verbatim}
\usepackage{amsmath}
\usepackage{amssymb}
\usepackage{booktabs}
\usepackage{multirow,multicol}

\usepackage{siunitx}
\usepackage[table]{xcolor}

\title{Hierarchical Mesh Transformers with Topology-Guided Pretraining for Morphometric Analysis of Brain Structures}

\author{
  Yujian Xiong \\
  Arizona State University \\
  \texttt{yxiong42@asu.edu} \\
  \And
  Mohammad Farazi \\
  Arizona State University \\
  \texttt{mfarazi@asu.edu} \\
  \And
  Yanxi Chen \\
  Arizona State University \\
  \texttt{ychen855@asu.edu} \\
  \And
  Wenhui Zhu \\
  Arizona State University \\
  \texttt{wzhu59@asu.edu} \\
  \And
  Xuanzhao Dong \\
  Arizona State University \\
  \texttt{xdong64@asu.edu} \\
  \And
  Natasha Lepore \\
  University of Southern California \\
  \texttt{nlepore@chla.usc.edu} \\
  \And
  Yi Su \\
  Banner Health \\
  \texttt{Yi.Su@bannerhealth.com} \\
  \And
  Raza Mushtaq \\
  Barrow Neurological Institute \\
  \texttt{raza.mushtaq@gmail.com} \\
  \And
  Stephen Foldes \\
  Barrow Neurological Institute \\
  \texttt{stephen.foldes@commonspirit.org} \\
  \And
  Andrew Yang \\
  Barrow Neurological Institute \\
  \texttt{Andrew.Yang@barrowbrainandspine.com} \\
  \And
  Yalin Wang\thanks{Corresponding author.} \\
  Arizona State University \\
  \texttt{ylwang@asu.edu} \\
}

\date{}

\renewcommand{\shorttitle}{\textit{arXiv} Template}

\hypersetup{
pdftitle={A template for the arxiv style},
pdfsubject={q-bio.NC, q-bio.QM},
pdfauthor={David S.~Hippocampus, Elias D.~Striatum},
pdfkeywords={First keyword, Second keyword, More},
}

\begin{document}
\maketitle

\begin{abstract}

Representation learning on large-scale unstructured volumetric and surface meshes poses significant challenges in neuroimaging, especially when models must incorporate diverse vertex-level morphometric descriptors—such as cortical thickness, curvature, sulcal depth, and myelin content—that carry subtle disease-related signals. Current approaches either ignore these clinically informative features or support only a single mesh topology, restricting their use across imaging pipelines. We introduce a hierarchical transformer framework designed for heterogeneous mesh analysis that operates on spatially adaptive tree partitions constructed from simplicial complexes of arbitrary order. This design accommodates both volumetric and surface discretizations within a single architecture, enabling efficient multi-scale attention without topology-specific modifications. A feature projection module maps variable-length per-vertex clinical descriptors into the spatial hierarchy, separating geometric structure from feature dimensionality and allowing seamless integration of different neuroimaging feature sets. Self-supervised pretraining via masked reconstruction of both coordinates and morphometric channels on large unlabeled cohorts yields a transferable encoder backbone applicable to diverse downstream tasks and mesh modalities. We validate our approach on Alzheimer's disease classification and amyloid burden prediction using volumetric brain meshes from ADNI, as well as focal cortical dysplasia detection on cortical surface meshes from the MELD dataset, achieving state-of-the-art results across all benchmarks.


\end{abstract}

\section{Introduction}

Learning from 3D medical meshes is a fundamental challenge in neuroimaging analysis. Structural MRI has become prominent for its non-invasive nature, high resolution, and suitability for longitudinal studies, playing a critical role in detecting fine-grained deformations such as cortical thinning or volumetric atrophy~\cite{qiu2009epidemiology}. Crucially, clinical pipelines routinely produce rich per-vertex morphometric signals including cortical thickness, curvature, sulcal depth, and myelin content, which carry critical diagnostic information for conditions such as Alzheimer's disease (AD) and focal cortical dysplasia (FCD).

Traditionally, MRI data are processed as voxel grids which have fixed resolution and are inherently limited in modeling intricate anatomical geometry~\cite{farazi2023tetcnn}, while unstructured meshes (tetrahedral/triangular) offer a topologically coherent and expressive alternative for both surface and interior anatomy. Yet existing frameworks are largely restricted to a single mesh type and operate on raw coordinates, ignoring the morphometric features that clinicians rely upon in practice.

\subsection{Related Work}

\noindent\textbf{Voxel and Graph-based Networks} extend 2D CNNs into 3D by discretizing space into regular grids~\cite{maturana2015voxnet,wu20153d}, but suffer from cubic computational cost. Sparse variants restrict computation to non-empty voxels via octrees~\cite{wang2017ocnn,riegler2017octnet} or hash tables~\cite{choy20194d}, and recent works further introduce windowed transformers over sparse voxels~\cite{wang2023dsvt,peng2024oa}. For irregular meshes, GNNs have become a natural choice~\cite{farazi2023tetcnn,farazi2023anisotropic,monti2017geometric}, yet most encode neighborhoods as topological graphs while overlooking underlying geometry, are limited to surface meshes~\cite{lahav2020meshwalker}, and none incorporate the per-vertex morphometric signals that are clinically important.

\noindent\textbf{Point-based Transformers} offer strong global modeling~\cite{zhao2021point,wu2024point} but face quadratic complexity on large meshes~\cite{cheng2022spherical}. Windowed attention~\cite{liu2021swin,farazi2024recipe} reduces cost but assumes regular density, causing information loss under spatially varying mesh structures~\cite{farazi2024recipe}. OctFormer~\cite{wang2023octformer} achieves near-linear complexity via adaptive octree windows, yet targets generic point clouds without heterogeneous mesh or morphometric feature support.

\noindent\textbf{Autoencoders} emerge as a powerful self-supervised pretraining paradigm, first demonstrated on images by MAE~\cite{he2022masked} and SimMIM~\cite{xie2022simmim}. This has been extended to 3D domains: Point-MAE~\cite{pang2022masked} pioneers masked modeling on point clouds with an asymmetric transformer, and Point-M2AE~\cite{zhang2022point} introduces multi-scale masking for hierarchical geometry learning. MAE principles have further been applied to LiDAR~\cite{min2023occupancy}, NeRF~\cite{irshad2024nerf}, and spatio-temporal data~\cite{wei2024t}. However, none of these works address pretraining on heterogeneous medical meshes with morphometric features, nor exploit shared mesh topology across subjects to amortize structural construction cost: leaving a clear gap for clinical neuroimaging.

We present \textbf{OctEncoder}, a unified octree transformer pretraining pipeline that addresses these gaps. Our key contributions are:
\begin{itemize}
    \item Multiple simplex-aware octrees construction which supports both tetrahedral and triangular meshes via octree-guided depthwise convolution.
    \item A geometry-morphometry fusion module enabling flexible per-vertex clinical feature integration without architectural changes.
    \item A MAE pretraining pipeline for general medical meshes, capturing both geometry features and vertex morphometry for any downstream tasks.
\end{itemize}

\section{Methods}

\subsection{Simplex-Aware Octree Construction}

\begin{figure}[t]
\centering
\includegraphics[width=0.98\textwidth]{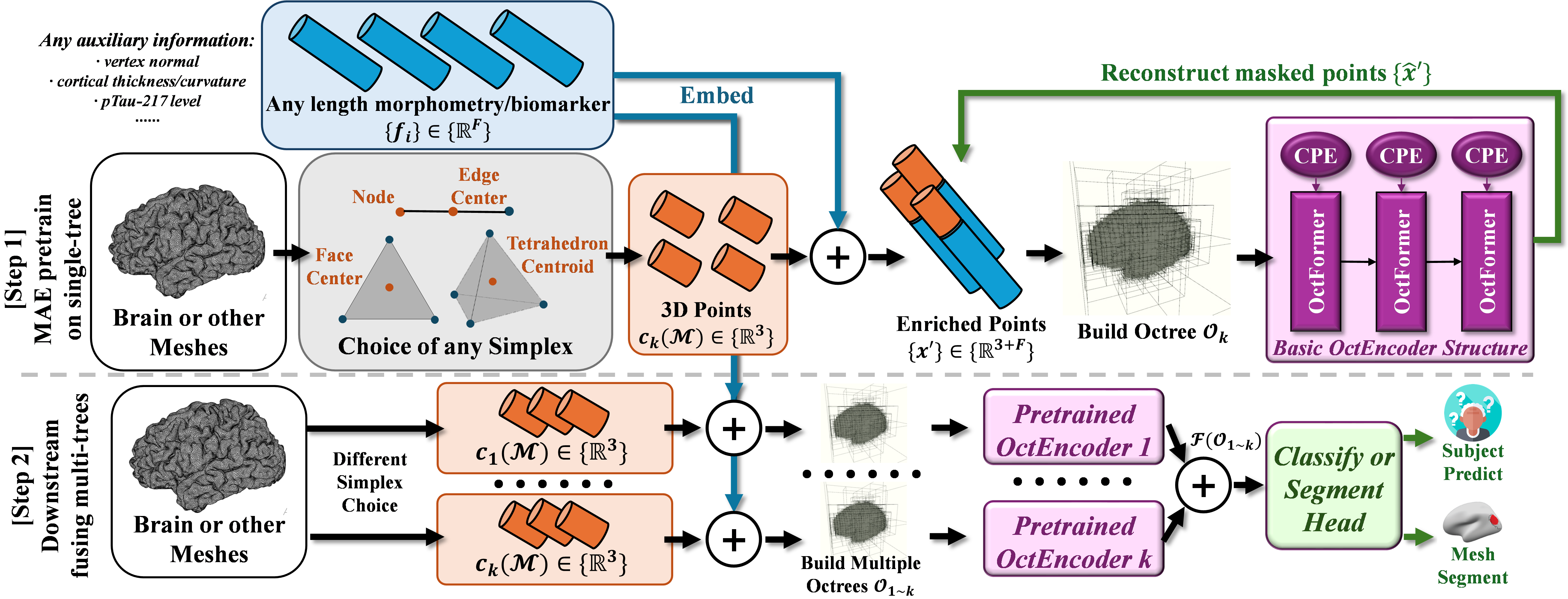} 
\caption{Overview of the framework on MAE (step-1) and downstream (step-2).}
\label{fig:model}
\end{figure}

As shown in Fig.~\ref{fig:model}, OctEncoder supports both tetrahedral and triangular meshes, with a flexible choice of representative points that can be adapted to dataset requirements. Formally, given a mesh $\mathcal{M} = (\mathcal{V}, \mathcal{S})$ where $\mathcal{V}$ is the vertex set and $\mathcal{S}$ the set of simplices, we define a representative point function $c: \mathcal{M} \rightarrow \{\mathbb{R}^3\}$ that produces a set of spatially localized points from the mesh.

For complex meshes, our pipeline can construct multiple complementary octrees. For example, the first uses tetrahedron centroids to capture volumetric interior geometry, while the second uses mesh vertices directly:
\[
c_1(\mathcal{M}) = \left\{ \frac{1}{4} \sum_{v_i \in t} v_i,\ t \in \mathcal{S} \right\}, \quad c_2(\mathcal{M}) = \left\{ v_i,\ v_i \in \mathcal{V} \right\}, \quad \dots \quad , \ c_K(\mathcal{M})
\]
The octree $\mathcal{O}_k$ is constructed by inserting all representative points $x \in c_k(\mathcal{M})$ up to a user-defined depth $d$, and nodes are ordered along a 3D Z-order space-filling curve for memory-contiguous window partitioning and efficient parallel construction~\cite{zhou2011data}. More generally, octrees can be constructed from any choice of simplex (vertex, edges/faces/tetrahedron center, etc.), and a learned weighted linear fusion strategy $\mathcal{F}(\mathcal{O}_1, \dots, \mathcal{O}_K)$ merges outputs across branches. The specific choice of simplex and number of octrees is a design decision driven by the dataset and task; our configuration reflects the setting used in our experiments.

Conditional Positional Encoding (CPE)~\cite{chu2021conditional} is applied independently within each octree branch before fusion, allowing each branch to develop spatially aware embeddings prior interaction.

\subsection{Auxiliary Clinical Feature Embedding}

Beyond constructing octrees from spatial coordinates alone, each vertex $v_i \in \mathcal{V}$ may carry a set of $F$ auxiliary feature channels derived from neuroimaging or geometry-morphometry pipelines, such as cortical thickness, curvature, etc. Given a per-vertex feature vector $\mathbf{f}_i \in \mathbb{R}^{F}$, any combination of per-vertex attributes can be incorporated, and the number of channels $F$ is flexible.

To embed such morphometric information into the octree representation, we augment each representative point $x \in c_k(\mathcal{M})$ by projecting the concatenation of its spatial coordinate and feature vector through a learnable linear layer:
\[
x' = W \begin{bmatrix} x \\ \mathbf{f}_i \end{bmatrix} + \mathbf{b}
\]
where $x' \in \mathbb{R}^{3+F}$ is the resulting enriched embedding. This allows OctEncoder to accommodate varying neuroimaging features without architectural changes.

\subsection{Octree Attention, MAE Pretraining and Downstream}

Transformer attention is computed within local windows defined by the octree's Z-order partitioning, keeping attention complexity near-linear in the number of nodes. Dilated attention complements local windows by sampling tokens at a fixed stride across the octree, enlarging the receptive field without additional memory cost. Each transformer block follows a standard residual design with layer normalization and a feedforward MLP~\cite{wang2023octformer}.

For pretraining on large unlabeled brain surfaces, we adopt a masked autoencoder strategy tailored to octree-structured data. A fixed proportion of octree tokens are masked, and the encoder processes only the visible subset to produce latent representations. A lightweight transformer decoder then reconstructs the masked tokens $\hat x' = [\hat x, \mathbf{\hat f}]$, supervised by a hybrid loss combining both Chamfer Distance over coordinates and MSE loss over morphometric features:
\[
\mathcal{L} = \underbrace{\sum_{p \in \{\hat x \}} \min_{q \in \{ x \}} \|p - q\|^2 + \sum_{q \in \{ x \} } \min_{p \in \{ \hat x \} } \|p - q\|^2}_{\mathcal{L}_{\text{chamfer}}} + \lambda \underbrace{\frac{1}{|\mathcal{M}|}\sum_{i \in \mathcal{M}} \|\hat{\mathbf{f}}_i - \mathbf{f}_i\|^2}_{\mathcal{L}_{\text{feat}}}
\]

After pretraining, the encoder serve as a general backbone that can be coupled with any task-specific head for downstream applications such as AD diagnosis or FCD segmentation. This flexibility allows OctEncoder to serve as a unified backbone for clinical neuroimaging tasks without retraining from scratch.


\section{Experimental Design}

\noindent\textbf{ADNI Classification:} We first pretrain our MAE encoder on the OASIS-3 dataset~\cite{lamontagne2019oasis}, which provides large-scale unlabeled sMRI scans across multiple sessions and subjects. The encoder is then used on the Alzheimer's Disease Neuroimaging Initiative (ADNI) dataset~\cite{jack2008alzheimer} for two downstream tasks, as summarized in Table~\ref{tab:datasets}: AD clinical diagnosis classification, and brain amyloid positivity (A$\beta$) prediction matched with PET scans and optional pTau-217 measurements.

\begin{figure}[t]
\centering
\includegraphics[width=0.98\textwidth]{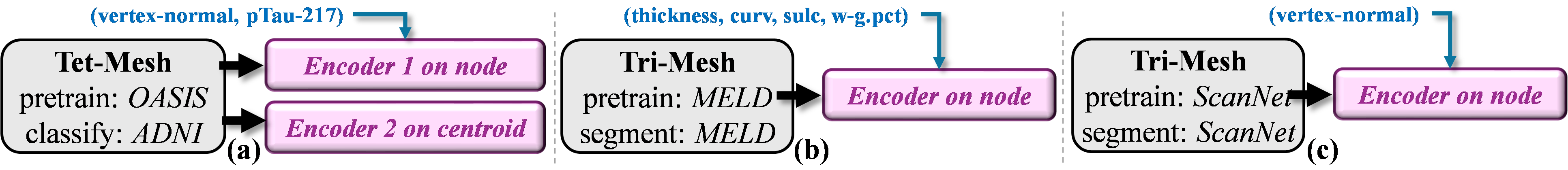} 
\caption{Overview of all 3 experiments and the auxiliary morphometries used.}
\label{fig:design}
\end{figure}

All MRIs are processed using FreeSurfer~\cite{fischl2012freesurfer} to reconstruct cortical surfaces, and volumetric tetrahedral meshes are generated via TetGen~\cite{hang2015tetgen} between pial and white-matter surfaces, producing approximately 130k--150k vertices per subject. Ground-truth A$\beta$ are derived from PET Centiloid values, where subjects with $\text{Centiloid} > 20$ are classified as A$\beta$ positive~\cite{su2013quantitative,klunk2015centiloid}. The pTau-217 biomarker, quantified via the PrecivityAD2 assay~\cite{eastwood2024precivityad2}, serves as an auxiliary biochemical label of A$\beta$ prediction~\cite{arranz2024diagnostic}. Details can be found in Fig~\ref{fig:design}.

For AD classification, we conduct pairwise binary tasks among AD/MCI/CN groups, comparing against tetrahedral-mesh baselines including ChebyNet~\cite{defferrard2016convolutional}, GAT~\cite{velivckovic2017graph}, and TetCNN~\cite{farazi2023tetcnn}. For amyloid prediction, we evaluate on mesh-only inputs and with auxiliary pTau-217 labels, comparing against logistic regression baselines using hippocampal volume and pTau-217. All models are evaluated using accuracy, sensitivity, and specificity.

\begin{table}[t]
    \centering
    \caption{Dataset and subjects summary for all 3 experiments.}
    \label{tab:datasets}
    \resizebox{\textwidth}{!}{
        \setlength{\tabcolsep}{6pt}
        \begin{tabular}{l | ccc | ccccccc}
            \toprule
            & \multicolumn{3}{c|}{\textbf{Pre-training}} & \multicolumn{7}{c}{\textbf{Downstream Tasks}} \\
            \midrule
            
            \rowcolor{gray!10} \multicolumn{11}{l}{\textbf{Experiment 1: ADNI \& OASIS (Tet-Mesh)}} \\
            \textbf{Dataset} & \multicolumn{3}{c|}{OASIS (Unlabeled)} & \multicolumn{3}{c}{ADNI (Predict AD)} & \multicolumn{4}{c}{ADNI (Predict Amyloid)} \\
            \cmidrule(lr){2-4} \cmidrule(lr){5-7} \cmidrule(l){8-11}
            Category & \multicolumn{3}{c|}{Total Samples} & AD & MCI & CN & \text{pTau-217} & Low & Mid & High \\
            \midrule
            Count & \multicolumn{3}{c|}{2,825} & 313 & 402 & 229 & A$\beta$+ & 19 & 70 & 265 \\
            & \multicolumn{3}{c|}{} & & & & A$\beta$$-$ & 335 & 96 & 23 \\
            
            
            \rowcolor{gray!10} \multicolumn{11}{l}{\textbf{Experiment 2 \& 3: ScanNet and MELD (Tri-Mesh)}} \\
            \textbf{Task} & \multicolumn{3}{c|}{\textbf{Scene Segmentation (ScanNet)}} & \multicolumn{7}{c}{\textbf{FCD Segmentation (MELD)}} \\
            \cmidrule(lr){2-4} \cmidrule(l){5-11}
            Stage & Pre-train & \multicolumn{2}{c|}{Downstream} & Pre-train & \multicolumn{6}{c}{Downstream} \\
            \midrule
            Data Type & Unlabeled & \multicolumn{2}{c|}{Labeled Scene} & Unlabeled & \multicolumn{6}{c}{CN \& Patients with FCD Labels} \\
            Count & 1,201 & \multicolumn{2}{c|}{1,513} & 942 & \multicolumn{6}{c}{373 CN \& 569 Patients} \\
            \bottomrule
        \end{tabular}
    }
\end{table}

\noindent\textbf{MELD Segmentation:} FCD is a leading cause of drug-resistant focal epilepsy, yet lesions are frequently MRI-negative and challenging to delineate precisely~\cite{ripart2025detection}. We evaluate OctEncoder on the publicly available MELD dataset~\cite{ripart2025detection}, a large multicenter cohort with cortical surface meshes processed via FreeSurfer to extract 34 per-vertex surface-based features including cortical thickness, gray-white matter intensity contrast, intrinsic curvature, sulcal depth, and FLAIR intensity sampled at multiple cortical depths. Each hemisphere is represented as a triangular surface mesh with 164k vertices and a corresponding binary lesion mask as the segmentation target. We adopt the published MELD Graph results~\cite{ripart2025detection} as our primary baseline. Performance is evaluated using subject-level sensitivity, specificity, and vertex-level lesion Intersection over Union (IoU).

\noindent\textbf{ScanNet Segmentation:} To test our generalizability beyond medical domains, we evaluate OctEncoder on large-scale 3D semantic segmentation using ScanNet~\cite{dai2017scannet}, comprising reconstructed indoor scenes annotated with 20 semantic categories. Only  face normals are used as auxiliary morphometries. Following standard protocol, training scenes are used for MAE pretraining and fine-tuning, with held-out scenes for validation and testing. We compare against OctFormer~\cite{wang2023octformer}, Point Transformer V1--V3~\cite{zhao2021point,wu2024point}, Mix3D~\cite{nekrasov2021mix3d}, O-CNN~\cite{wang2017ocnn}, and TTT-KD~\cite{weijler2024ttt}, evaluated using mean Intersection-over-Union (mIoU).

\section{Results}

\begin{figure}[t]
\centering
\includegraphics[width=0.98\textwidth]{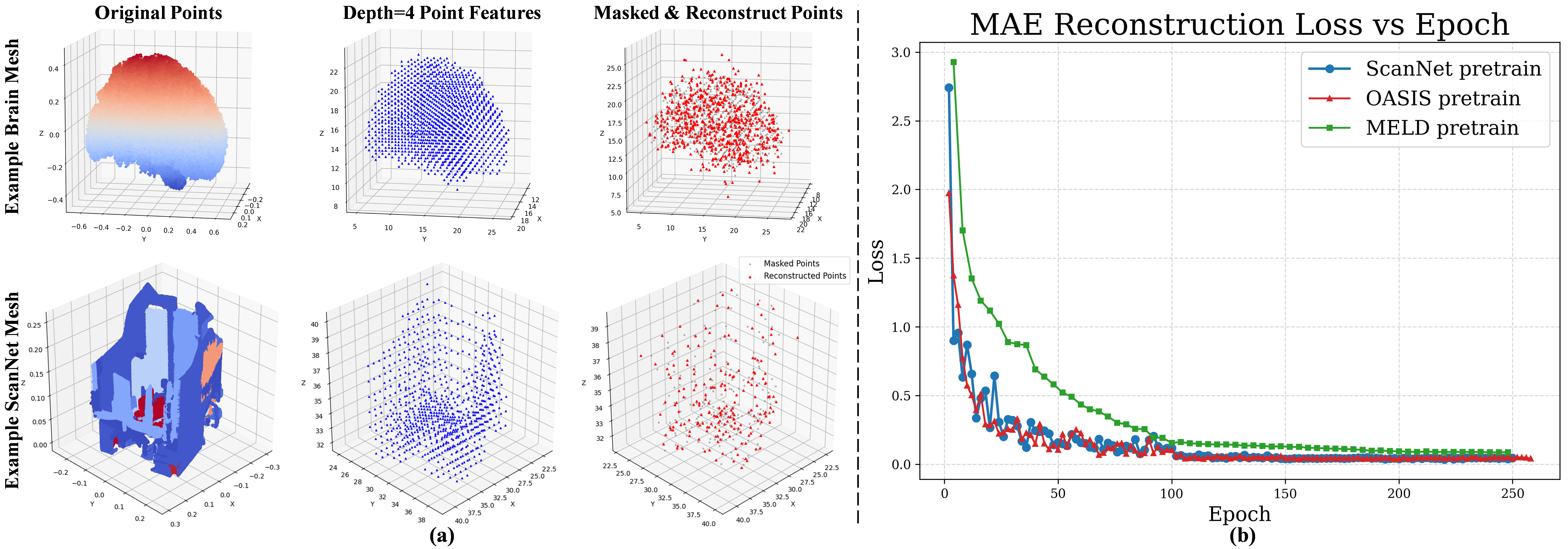}
\caption{\textbf{(a)} Visualization of MAE pretraining on brain tet-meshes (top) and ScanNet tri-meshes (bottom). From left: original points, octree depth 4 point features, and masked input (\textcolor{gray}{gray}) with reconstruction (\textcolor{red}{red}). \textbf{(b)} MAE reconstruction loss curves for OASIS (red), ScanNet (blue) and MELD (green) pretraining.}
\label{fig:mae}
\end{figure}

\noindent\textbf{MAE Pretraining Performance.} As shown in Figure~\ref{fig:mae}, OctEncoder learns high-quality geometric representations across both mesh domains, with reconstructed points closely matching masked regions even under aggressive masking. Both pretraining curves converge steadily within roughly 100 epochs, after which additional epochs contribute negligible reconstruction improvement.

\noindent\textbf{Computation Costs.} All brain experiments are run on a single NVIDIA Quadro RTX 5000 GPU, where MAE pretraining takes approximately 1.5 GPU hours and downstream fine-tuning completes in around 2 hours. ScanNet experiments are run on 4 identical GPUs, with MAE pretraining requiring approximately 20 GPU hours and segmentation training approximately 33 GPU hours.


\begin{table}[t]
    \centering
    \caption{ADNI Tet-mesh Results: Classification and Amyloid Prediction.}
    \label{tab:adni_results}
     
    \resizebox{\textwidth}{!}{
        
        \setlength{\tabcolsep}{6pt} 
        \begin{tabular}{l ccc ccc ccc ccc}
            \toprule
            & \multicolumn{9}{c}{\textbf{Alzheimer's Disease Classification}} & \multicolumn{3}{c}{\textbf{Amyloid Prediction}} \\
            \cmidrule(lr){2-10} \cmidrule(l){11-13}
            & \multicolumn{3}{c}{AD vs CN} & \multicolumn{3}{c}{AD vs MCI} & \multicolumn{3}{c}{MCI vs CN} & \multicolumn{3}{c}{Medium-Risk Group} \\
            \cmidrule(lr){2-4} \cmidrule(lr){5-7} \cmidrule(lr){8-10} \cmidrule(l){11-13}
            \textbf{Model} & ACC & SEN & SPE & ACC & SEN & SPE & ACC & SEN & SPE & ACC & SEN & SPE \\
            \midrule
            \rowcolor{gray!10} \multicolumn{13}{l}{\textit{Baselines only for Amyloid Prediction}} \\
            LR on Hippo-Vol. & --- & --- & --- & --- & --- & --- & --- & --- & --- & 0.450 & 0.529 & 0.391 \\
            LR on pTau-217   & --- & --- & --- & --- & --- & --- & --- & --- & --- & 0.675 & 0.750 & 0.625 \\
            LR on Hippo + pTau   & --- & --- & --- & --- & --- & --- & --- & --- & --- & 0.675 & 0.750 & 0.625 \\
            \midrule
            \rowcolor{gray!10} \multicolumn{13}{l}{\textit{Tetrahedron Mesh GNN Methods}} \\
            ChebyNet & 0.870 & 0.881 & 0.850 & 0.703 & \textbf{0.790} & 0.616 & \underline{0.735} & \textbf{0.778} & 0.667 & 0.677 & 0.563 & 0.800 \\
            GAT      & 0.858 & 0.873 & 0.836 & \underline{0.727} & 0.630 & \underline{0.773} & 0.722 & 0.763 & 0.660 & 0.677 & 0.611 & 0.769 \\
            TetCNN   & \underline{0.876} & \underline{0.886} & \underline{0.859} & 0.709 & \underline{0.660} & 0.769 & 0.730 & 0.761 & \underline{0.700} & 0.690 & 0.684 & 0.694 \\
            \midrule
            \rowcolor{blue!5} \textbf{Ours} & & & & & & & & & & & & \\
            \textbf{OctEncoder} & \textbf{0.907} & \textbf{0.902} & \textbf{0.914} & \textbf{0.731} & 0.650 & \textbf{0.812} & \textbf{0.782} & \underline{0.761} & \textbf{0.807} & \underline{0.763} & \underline{0.751} & \underline{0.774} \\
            \textbf{OctEncoder + pTau} & --- & --- & --- & --- & --- & --- & --- & --- & --- & \textbf{0.815} & \textbf{0.781} & \textbf{0.848} \\
            \bottomrule
        \end{tabular}
    }
\end{table}

\noindent\textbf{AD Classification.} OctEncoder consistently outperforms all baselines across all three pairwise tasks (Table~\ref{tab:adni_results}). For AD vs. CN, the model achieves strong accuracy and well-balanced sensitivity and specificity, confirming robust discrimination between late-stage Alzheimer's and healthy controls. The most clinically significant result is in MCI vs. CN, where OctEncoder improves accuracy by 4.7\% over the second-best model, demonstrating superior sensitivity to early and subtle pathological differences in prodromal subjects: a task of direct relevance for identifying individuals at risk of progression to AD. In AD vs. MCI, ChebyNet achieves slightly higher sensitivity, but at the cost of substantially lower specificity, reflecting a tendency to over-predict positives rather than reliably discriminate between groups.

\noindent\textbf{Amyloid Positivity Prediction.} We focus on the medium-risk subgroup, where amyloid status is most ambiguous and biomarker-based prediction is most clinically uncertain. Notably, adding hippocampal volume to pTau-217 in logistic regression yields no improvement, suggesting these two conventional features are largely redundant in this subgroup. OctEncoder alone surpasses all biomarker-only baselines with balanced sensitivity and specificity, demonstrating that tetrahedral mesh geometry captures pathological signals not reflected in scalar biomarkers. Its fusion with pTau-217 achieves the strongest overall performance, with accuracy of 0.815, confirming that structural mesh features and blood-based biomarkers carry complementary information for identifying amyloid pathology in clinically ambiguous cases.

\begin{table}[t]
    \centering
    \caption{Segmentation results on ScanNet Scene and MELD FCD segmentation.}
    \label{tab:segment}
    \resizebox{\textwidth}{!}{
        \setlength{\tabcolsep}{6pt}
        \begin{tabular}{l ccc cccc >{\columncolor{blue!5}}c} %
            \toprule
            
            \textbf{Method} & \multicolumn{4}{c}{MELD Graph Neural Network~\cite{ripart2025detection}} & \multicolumn{3}{c}{OctFormer~\cite{wang2023octformer}} & \textbf{OctEncoder} \\
            \cmidrule(r){2-5} \cmidrule(l){6-8} \cmidrule(l){9-9}
            \rowcolor{gray!10} \multicolumn{9}{l}{\textit{MELD Segmentation Performance}} \\
            Lesion IoU & \multicolumn{4}{c}{0.30} & \multicolumn{3}{c}{\underline{0.34}} & \textbf{0.51} \\
            Subject Sensitivity & \multicolumn{4}{c}{0.70} & \multicolumn{3}{c}{\underline{0.73}} & \textbf{0.78} \\
            Subject Specificity & \multicolumn{4}{c}{\underline{0.60}} & \multicolumn{3}{c}{0.52} & \textbf{0.63} \\
            
            \midrule
            \textbf{Method} & PT & Mix3D & O-CNN & PT-V2 & OctFormer & PT-V3 & TTT-KD & \textbf{OctEncoder} \\
            \midrule
            \rowcolor{gray!10} \multicolumn{9}{l}{\textit{ScanNet Segmentation Performance}} \\
            Mean IoU & 0.706 & 0.736 & 0.745 & 0.754 & 0.757 & 0.775 & \underline{0.776} & \textbf{0.777} \\
            
            \bottomrule
        \end{tabular}
    }
\end{table}

\noindent\textbf{FCD Lesion Segmentation.} OctEncoder substantially outperforms the published MELD Graph baseline~\cite{ripart2025detection} across all three metrics (Table~\ref{tab:segment}). The most striking improvement is in lesion IoU, which increases from 0.30 to 0.51, reflecting significantly more precise delineation of dysplastic cortical regions. Subject-level sensitivity also improves from 0.70 to 0.78, indicating that OctEncoder detects a greater proportion of patients with FCD. Specificity improves more modestly, suggesting the primary gain is in lesion localization accuracy rather than false positive reduction. These results demonstrate that the octree transformer architecture, equipped with rich per-vertex morphometric features and MAE pretraining on unlabeled cortical meshes, is well-suited to the subtle and spatially irregular patterns characteristic of FCD.

\noindent\textbf{3D Semantic Segmentation.} OctEncoder achieves the highest mIoU among all compared methods on ScanNet, marginally surpassing TTT-KD and Point Transformer V3 while offering substantially reduced computational cost through octree-guided encoding and MAE pretraining. The consistent improvement over earlier transformer-based models including OctFormer and Point Transformer V1/V2 confirms the benefit of combining hierarchical octree partitioning with masked pretraining for dense semantic segmentation in large-scale indoor scenes.

\subsection{Ablation Studies}

\begin{table}[t]
    \centering
    \caption{Ablation study on ADNI AD vs. CN task, evaluating 4 different components.}
    \label{tab:ablation}
    \resizebox{0.9\textwidth}{!}{ 
        \setlength{\tabcolsep}{12pt} 
        \begin{tabular}{lccc}
            \toprule
            \textbf{Method} & ACC & SEN & SPE \\
            \midrule

            \rowcolor{gray!10} \multicolumn{4}{l}{\textit{Positional Encoding}} \\
            No Positional Encoding & 0.863 & 0.842 & 0.884 \\
            \rowcolor{blue!5} + CPE \textbf{(Proposed)} & \textbf{0.907} & \textbf{0.902} & \underline{0.914} \\
            + RPE & 0.882 & 0.873 & 0.890 \\
            
            \midrule
            
            \rowcolor{gray!10} \multicolumn{4}{l}{\textit{Octree Ordering}} \\
            \rowcolor{blue!5} Z-order Curve \textbf{(Proposed)} & \textbf{0.907} & \textbf{0.902} & \underline{0.914} \\
            Hilbert Curve & \underline{0.899} & \underline{0.876} & \textbf{0.922} \\
            
            \midrule
             
            \rowcolor{gray!10} \multicolumn{4}{l}{\textit{Simplex Fusion Strategy}} \\
            $\mathcal{O}_1$ on nodes only & 0.876 & 0.851 & 0.901 \\
            \rowcolor{blue!5} $\mathcal{O}_1$ on nodes + $\mathcal{O}_2$ on centroid \textbf{(Proposed)} & \textbf{0.907} & \textbf{0.902} & \underline{0.914} \\
            
            \midrule
            
            \rowcolor{gray!10} \multicolumn{4}{l}{\textit{MAE Pretraining}} \\
            No MAE pretrain & 0.790 & 0.768 & 0.809 \\
            \rowcolor{blue!5} + MAE pretrain \textbf{(Proposed)} & \textbf{0.907} & \textbf{0.902} & \underline{0.914} \\
            
            \bottomrule
        \end{tabular}
    }
\end{table}

Due to page constraints, ablation studies are reported on the AD vs. CN task only. As Table~\ref{tab:ablation}, each proposed component contributes meaningfully to final performance. CPE yields the largest positional encoding gain over both the no-encoding baseline and conventional RPE, confirming the benefit of learning position-dependent features adaptively within the octree structure. For octree ordering, Z-order and Hilbert curves perform comparably, but Z-order is preferred for its better computational efficiency at scale. Fusing node and tetrahedron-center octrees $(\mathcal{O}_1,\mathcal{O}_2)$ improves over the single-octree variant, demonstrating the value of multi-view geometric aggregation. Most critically, removing MAE pretraining causes the largest single performance drop across all ablations, confirming that self-supervised pretraining on large unlabeled meshes is foundational rather than merely supplementary for high-quality mesh representation learning.

\section{Conclusion}


We present OctEncoder, a unified octree transformer for heterogeneous medical mesh analysis. Beyond OctFormer, our key advances are: simplex-aware multi-tree construction supporting both tetrahedral and triangular meshes via flexible simplicial complex fusion; a geometry-morphometry embedding that integrates arbitrary per-vertex clinical features without architectural changes; and a general MAE pretraining pipeline that jointly reconstructs geometry and morphometry across any mesh modality, exploiting shared topology to reduce pretraining cost.

Validated across three clinically distinct tasks on ADNI tetrahedral brain meshes, MELD cortical surface meshes, and ScanNet indoor scenes, OctEncoder achieves state-of-the-art performance in all settings, confirming that unified geometry-morphometry encoding with self-supervised pretraining is both clinically effective and broadly generalizable.

\bibliographystyle{unsrtnat}
\bibliography{main}  






\end{document}